\documentclass[10pt,twocolumn,letterpaper]{article}

\usepackage{cvpr}
\usepackage{times}
\usepackage{epsfig}
\usepackage{graphicx}
\usepackage{amsmath}
\usepackage{amssymb}

\usepackage{epsfig}
\usepackage{amsmath}
\usepackage{amssymb}
\usepackage{color}
\usepackage{comment}
\usepackage{algorithm}
\usepackage{algorithmic}

\usepackage{enumitem}
\usepackage{amsfonts}
\usepackage{extarrows}
\usepackage{multirow}
\usepackage{rotating}

\usepackage[pagebackref=true,breaklinks=true,letterpaper=true,colorlinks,bookmarks=false]{hyperref}

\cvprfinalcopy 

\def\cvprPaperID{2145} 

\ifcvprfinal\fi
\begin{document}

\title{Network Transplanting}

\author{Quanshi Zhang$^{a}$, Yu Yang$^{b}$, Qian Yu$^{c}$, Ying Nian Wu$^{b}$\\
$^{a}$Shanghai Jiao Tong University\\
$^{b}$University of California, Los Angeles\\
$^{c}$University of California, Berkeley\\
}

\maketitle

\begin{abstract}
This paper focuses on a new task, \emph{i.e.} transplanting a category-and-task-specific neural network to a generic, modular network without strong supervision. We design an functionally interpretable structure for the generic network. Like building LEGO blocks, we teach the generic network a new category by directly transplanting the module corresponding to the category from a pre-trained network with a few or even without sample annotations. Our method incrementally adds new categories to the generic network but does not affect representations of existing categories. In this way, our method breaks the typical bottleneck of learning a net for massive tasks and categories, \emph{i.e.} the requirement of collecting samples for all tasks and categories at the same time before the learning begins. Thus, we use a new distillation algorithm, namely back-distillation, to overcome specific challenges of network transplanting. Our method without training samples even outperformed the baseline with 100 training samples.
\end{abstract}


\section{Introduction}

Besides end-to-end learning a black-box neural network, in this paper, we propose a new deep-learning methodology, \emph{i.e.} network transplanting.
Instead of learning from scratch, network transplanting aims to merge several convolutional networks that are pre-trained for different categories and tasks to build a generic, distributed neural network.

Network transplanting is of special values in both theory and practice. We briefly introduce key deep-learning problems that network transplanting deals with as follows.

\subsection{Future potential of learning a universal net}
\label{sec:universal}

Instead of learning different networks for different applications, building a universal net with a compact structure for various categories and tasks is one of ultimate objectives of AI. In spite of the gap between current algorithms and the target of learning a huge universal net, it is still meaningful for scientific explorations along this direction. Here, we list key issues of learning a universal net, which are not commonly discussed in the current literature of deep learning.

\noindent
$\bullet\;$\textbf{The start-up cost \emph{w.r.t.} sample collection} is also important, besides the total number of training annotations. Traditional methods usually require people to simultaneously prepare training samples for all pairs of categories and tasks before the learning begins. However, it is usually unaffordable, when there is a large number of categories and tasks. In comparison, our method enables a neural network to sequentially absorb network modules of different categories one-by-one, so the algorithm can start without all data.

\noindent
$\bullet\;$\textbf{Massive distributed learning \& weak centralized learning:} Distributing the massive computation of learning the network into local computation centers all over the world is of great practical values. There exist numerous networks locally pre-trained for specific tasks and categories in the world. Centralized network transplanting physically merges these networks into a compact universal net with a few or even without any training samples.

\noindent
$\bullet\;$\textbf{Delivering models or data:} Our delivering pre-trained networks to the computation center is usually much cheaper than collecting and sending raw training data in practice.

\begin{figure*}[t]
\centering
\includegraphics[width=0.9\linewidth]{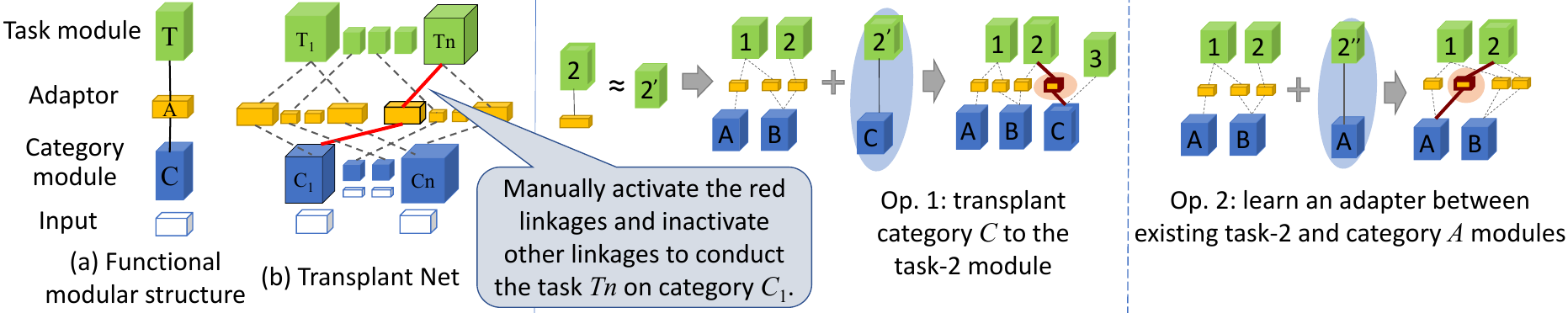}
\caption{Building a transplant net. We propose a theoretical solution to incrementally merging category modules from teacher nets into a transplant (student) net with a few or without sample annotations. The transplant net has an interpretable, modular structure. A category module, \emph{e.g.} a cat module, provides cat features to different task modules. A task module, \emph{e.g.} a segmentation module, serves for various categories. We show two typical operations to learn transplant nets\textcolor{red}{{\protect\footnotemark[1]}}. Blue ellipses show modules in teacher nets used for transplant. Red ellipses indicate new modules added to the transplant net. Unrelated adapters in each step are omitted for clarity.}
\label{fig:problem}
\end{figure*}

\noindent
$\bullet\;$\textbf{Middle-to-end semantic manipulation for application:} How to efficiently organize and use the knowledge in the net is also a crucial problem. We use different modules in the network to encode knowledge of different categories and that of different tasks. Like building LEGO blocks, people can manually connect a category module and a task module to accomplish a certain application (see Fig.~\ref{fig:problem}(left)).


\begin{table*}
\begin{center}
\resizebox{\linewidth}{!}{
\begin{tabular}{p{2.8cm}|p{3.5cm}p{4.5cm}cccc}
& \multicolumn{1}{|c}{Annotation cost} & \multicolumn{1}{c}{Sample preparation} & \!\!Interpretability\!\! & Catastrophic forgetting & \!\!{\small Modular manipulate}\!\! & Optimization\\
\hline
Directly learning a multi-task net & Massive & Simultaneously prepare samples for all tasks and categories & Low & -- & Not support & back prop.\\
\hline
Transfer- / meta- / continual-learning & Some support weakly-supervised learning & Some learn a category/task after another & Usually low & \!\!Most algorithmically alleviate\!\! & No support & back prop.\\
\hline
Transplanting & A few or w/o annotations & Learns a category after another & High & Physically avoid & Support & \!\!back-back prop.\!\!\\
\hline
\end{tabular}}
\vspace{2pt}
\caption{Comparison between network transplanting and other studies. Note that this table can only summarize mainstreams in different research directions considering the huge research diversity. Please see Section~\ref{sec:related} for detailed discussions of related work.}
\label{tab:diff}
\end{center}
\end{table*}

\subsection{Task of network transplanting}

To solve above issues, we propose network transplanting, \emph{i.e.} building a generic model by gradually absorbing networks locally pre-trained for specific categories and tasks. We design an interpretable modular structure for a target network, namely a \textit{transplant net}, where each module is functionally meaningful. As shown in Fig.~\ref{fig:problem}(left), the transplant net consists of three types of modules, \emph{i.e.} category modules, task modules, and adapters. Each category module extracts general features for a specific category (\emph{e.g.} the dog). Each task module is learned for a certain task (\emph{e.g.} classification or segmentation) and is shared by different categories. Each adapter projects output features of a category module to the input space of a task module. Each category/task module is shared by multiple tasks/categories.

We can learn an initial transplant net with very few tasks and categories in the scenario of traditional multi-task/category learning. Then, we gradually grow the transplant net to deal with more categories and tasks via network transplanting. Network transplanting can be conducted with or without human annotations as additional supervision. We summarize two typical types of transplanting operations in Fig.~\ref{fig:problem}(right). The core technique is to learn an adapter to connect a task module in the transplant net and a category module from another network\footnote[1]{Please see supplementary materials for details.}.

\begin{figure*}[t]
\centering
\includegraphics[width=0.86\linewidth]{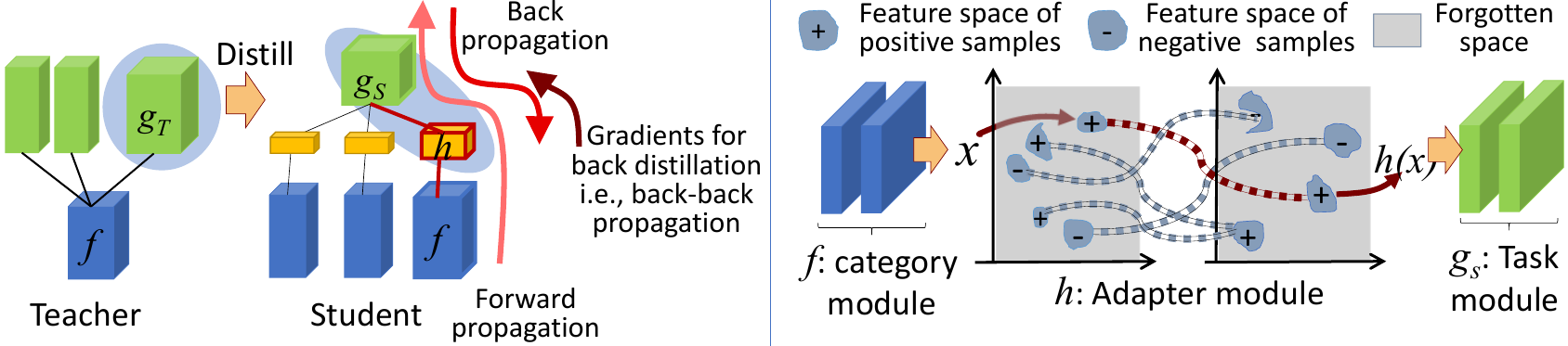}
\caption{Overview. (left) Given a teacher net and a student net, we aim to learn an adapter $h$ via distillation. The teacher net has a category module $f$ for a single or multiple tasks. The student net contains a task module {\small$g_{S}$} for other categories. We transplant $f$ to the student net by using $h$ to connect $f$ and {\small$g_{S}$}, in order to enable the task module {\small$g_{S}$} to deal with the new category $f$. As shown in green ellipses, our method distills knowledge from {\small$g_{T}$} of a teacher net to the {\small$h \circ g_{S}$} modules of the student net. Three red curves show directions of forward propagation, back propagation, and gradient propagation of back distillation. (right) During the transplant, the adapter $h$ learns potential projections between $f$'s output feature space and {\small$g_{S}$}'s input feature space.}
\label{fig:task}
\end{figure*}

The elementary transplanting operation is shown in Fig.~\ref{fig:task}(left). We are given a transplant net with a task module $g_{S}$ that is learned to accomplish a certain task for many categories, except for the category $c$. We hope the task module $g_{S}$ to deal with the new category $c$, so we need another network (namely, a \textit{teacher net}) with a category module $f$ and a task module $g_{T}$. The teacher net is pre-trained for the same task on the category $c$. We may (or may not) have a few training annotations of category $c$ for the task. Our goal is to transplant the category module $f$ in the teacher net to the transplant net.

Note that we just learn a small adapter module to connect $f$ to $g_{S}$. We do \textbf{not} fine-tune $f$ and $g_{S}$ during the transplanting process to avoid damaging their generality.

However, learning adapters but fixing parameters of category and task modules proposes specific challenges to deep-learning algorithms. Therefore, in this study, we proposed a new algorithm, namely \textit{back distillation}, to overcome these challenges. The back-distillation algorithm uses the cascaded modules of the adapter and $g_{S}$ to mimic upper layers of the pre-trained teacher net. This algorithm requires the transplant net to have similar gradients/Jacobian with the teacher net \emph{w.r.t.} $f$'s output features for distillation. In experiments, our back-distillation method without any training samples even outperformed the baseline with 100 training samples (see Table~\ref{tab:exp1}(left)).

\subsubsection{Difference to previous knowledge transferring}


Although most transfer-learning algorithms cannot be directly used to solve core problems mentioned in Section~\ref{sec:universal}, the proposed network transplanting is close to the spirit of continual learning (or lifelong learning)~\cite{ProgressiveNN,PathNet,continualLearning,LifelongLearning,metaContinual}. As an exploratory research, we summarize our essential differences from traditional studies in Table~\ref{tab:diff}.

\noindent
$\bullet\;$\textbf{Modular interpretability $\rightarrow$ more controllability:} Besides the discrimination power, the interpretability is another important property of a neural network, which has received increasing attention in recent years. In particular, traditional transfer learning and knowledge distillation that are implemented in a black-box manner~\cite{ProgressiveNN,netMerger}, so the generalization process requires careful control. Whereas, our transplant net clarifies the functional meaning of each intermediate network module, which makes the knowledge-transferring process more controllable. \emph{I.e.} the interpretable structure clearly points out network modules that are related to the target application.

\noindent
$\bullet\;$\textbf{Bottleneck of transferring upper modules\textcolor{red}{\footnotemark[1]}:} Most deep-learning strategies are not suitable to directly transfer pre-trained upper modules, including (i) directly optimizing the task loss on the top of the network and (ii) traditional distillation methods~\cite{distill}. These methods mainly transfer low-layer features and learn new upper modules to reuse these features, rather than directly transfer pre-trained upper modules. In contrast, network transplanting just allows us to modify the lower adapter, when we transfer a pre-trained upper task module $g_{S}$ to a new category. It is not permitted to modify the upper module. This requirement physically avoids the catastrophic forgetting, but it is difficult to optimize a lower adapter if the upper $g_{S}$ is fixed (see Section~\ref{sec:challenge} for theoretical analysis). Meanwhile, it is difficult to distill knowledge from the teacher net to the adapter. It is because except for the final network output, high-layer features in $g_{S}$ and those in the teacher net are not semantically aligned.

Thus, in this paper, the proposed back distillation first breaks the bottleneck of transferring upper modules.

\noindent
$\bullet\;$\textbf{Catastrophic forgetting:} Continually learning new jobs without hurting the performance of old jobs is a key issue for continual learning~\cite{ProgressiveNN,continualLearning}. Our method exclusively learns the adapter to physically prevent the learning of new categories from changing existing modules. Furthermore, when a transplant net has been constructed, we can optionally fine-tune a task/category module based on different categories/tasks to ensure the network generality.

\subsubsection{Summarization}

We can summarize contributions of this study as follows. (i) We propose a new deep-learning method, network transplanting with a few or even without additional training annotations, which can be considered as a theoretical solution to three issues in Section~\ref{sec:universal}. (ii) We develop an optimization algorithm, \emph{i.e.} back-distillation, to overcome specific challenges of network transplanting. (iii) Preliminary experiments proved the effectiveness of our method. Our method significantly outperformed baselines.


\section{Related work}
\label{sec:related}

Because network transplanting is a new concept in machine learning, we would like to discuss its connections to different state-of-the-art algorithms. Firstly, we propose a new modular structure for networks, which disentangles a black-box network into different meaningful modules. Similarly, some studies have explored new representation structures instead of neural networks, such as forests and decision trees~\cite{deepForest1,deepForest2,distillDecisionTree,RNNTree} and automatic learning of optimal network structures~\cite{LearnNetStruct1,LearnNetStruct2,LearnNetStruct3,LearnNetStruct4}. \cite{NetMixOfExperts} learned a large modular neural network with thousands of sub-networks.

\textbf{Interpretability:} Unlike above studies, our dividing a network into functionally meaningful modules makes the structure interpretable. Other studies of enhancing network interpretability mainly either learn disentangled features filters/capsules in middle layers~\cite{interpretableCNN,InterRCNN,capsule} or learn meaningful input codes of generative nets~\cite{infoGAN,betaVAE}.

\begin{figure*}[t]
\centering
\includegraphics[width=0.9\linewidth]{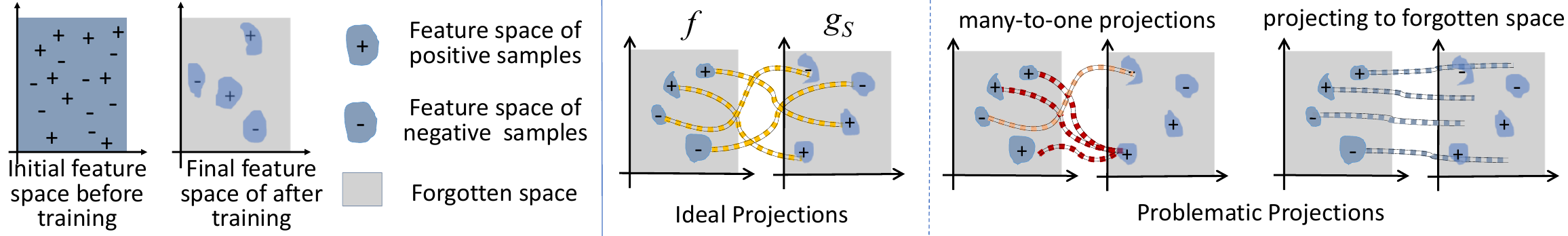}
\vspace{2pt}
\caption{Feature space of a middle layer. (left) When we initialize parameters of a CNN, middle-layer features randomly cover all area in the feature space. The learning process forces the CNN to focus on typical feature spaces of samples and produces vast forgotten space. (right) We illustrate three toy examples of space projection that are estimated by the adapter.}
\label{fig:space}
\end{figure*}

\textbf{Meta-learning \& transfer learning:} Meta-learning~\cite{meta1,meta2,meta3} aims to extract generic knowledge shared by different tasks/categories/models to guide the learning of a specific model. Transfer-learning methods transfer network knowledge through categories~\cite{CNNAnalysis_2} or datasets~\cite{UnsuperTransferCNN}. Especially, continual learning~\cite{ProgressiveNN,PathNet,continualLearning,LifelongLearning,metaContinual} transfers knowledge from previous tasks to guide the new task. \cite{ProgressiveNN,LifelongLearning} expanded the network structure during the learning process. In contrast, our study defines modular network structures with strict semantics, which allows people to semantically control the knowledge to transfer. Meanwhile, network transplanting physically avoids the catastrophic forgetting. In addition, our back distillation method solves the challenge of transferring upper modules, which is different from traditional transferring of low-layer features.

\textbf{Distillation:} \cite{distill} proposed the knowledge distillation to transfer knowledge between networks. Some recent studies~\cite{distillDecisionTree,RNNTree} distilled network knowledge into decision trees. \cite{distillDistributed} proposed an online distillation method to efficiently learn distributed networks. \cite{distillAttention,distillJacobian} distilled the attention distribution/Jacobian from the teacher network to the student network, which is related to our back-distillation technique. However, these Jacobian distillation methods are hampered considering specific challenges of network transplanting (see Section~\ref{sec:backDistill} and the appendix for details). To overcome these challenges, in this study, we design a new method of back-distillation, which uses pseudo-gradients instead of using real gradients for distillation. We also balance magnitudes of neural activations between two networks, which is necessary for network transplanting.

\section{Algorithm of network transplanting}

\textbf{Overview:} As shown in Fig.~\ref{fig:task}(left), we are given a teacher net for a single or multiple tasks \emph{w.r.t.} a certain category $c$. Let the category module $f$ in the bottom of the teacher net have $m$ layers, and it connects a specific task module {\small$g_{T}$} in upper layers. We are also given a transplant net with a generic task module {\small$g_{S}$}, which has been learned for multiple categories except for the category $c$.

The initial transplant net with a task module {\small$g_{S}$} (before transplanting) can be learned via traditional scenario of learning from samples of some categories. We can roughly regard {\small$g_{S}$} to encode generic representations for the task. Similarly, the category module $f$ extracts generic features for multiple tasks. Thus, we do not fine-tune {\small$g_{S}$} or $f$ to avoid decreasing their generality.

Our goal is to transplant $f$ to {\small$g_{S}$} by learning an adapter $h$ with parameters $\theta_{h}$, so that the task module {\small$g_{S}$} can deal with the new category module $f$.

The basic idea of network transplanting is that \textit{we use the cascaded modules of $h$ and {\small$g_{S}$} to mimic the specific task module {\small$g_{T}$} in the teacher net.} We call the transplant net a \textit{student net}. Let $x$ denote the output feature of the category module $f$ given an image $I$, \emph{i.e.} {\small$x=f(I)$}. {\small$y_{T}$} and {\small$y_{S}$} are given as outputs of the teacher net and the student net, respectively. Thus, network transplanting can be described as
\begin{small}
\begin{equation}
\!y_{T}\!=\!g_{T}(x),\; y_{S}\!=\!g_{S}(h(x)), \; y_{T}\!\approx\!y_{S} \;\Rightarrow\; g_{S}(h(\cdot))\!\approx\!g_{T}(\cdot)\!\!\!
\end{equation}
\end{small}


\subsection{Problem of space projection \& back-distillation}
\label{sec:challenge}

It is a challenge to let an adapter $h$ project the output feature space of $f$ properly to the input feature space of {\small$g_{S}$}. The information bottleneck theory~\cite{InformationBottleneck,InformationBottleneck2} shows that a network selectively \textit{forgets} certain space of middle-layer features and gradually focuses on discriminative features during the learning process (see Fig.~\ref{fig:space}(left)). Thus, both the output of $f$ and the input of {\small$g_{S}$} have vast \textit{forgotten space}. Features in the forgotten input space of {\small$g_{S}$} cannot pass most feature information through ReLU layers in {\small$g_{S}$} and reach {\small$y_{S}$}. The forgotten output space of $f$ is referred to the space that does not contains $f$'s output features.

Vast forgotten feature spaces significantly boost difficulties of learning. Since valid input features of {\small$g_{S}$} usually lie in low-dimensional manifolds, most features of the adapter fall inside the forgotten space. \emph{I.e.} $g_{S}$ will not pass most information of input features to network outputs. Consequently, the adapter will not receive informative gradients of the loss for learning.

Fig.~\ref{fig:space}(right) illustrates ideal and problematic projections. A typical problematic projection is to project a feature $x$ to a forgotten input space of {\small$g_{S}$}. Another typical problem is many-to-one projections, which limit the diversity of features and decrease the representation capability of the student net. More crucially, initial many-to-one projections significantly affect the further learning process, because the back-propagation takes current space projections as anchors to fine-tune the network.

To learn good projections, we propose to force the gradient (also known as attention, Jacobian) of the student net to approximate that of the teacher, which is a necessary condition of {\small$g_{S}(h(\cdot))\!\approx\!g_{T}(\cdot)$}.
\begin{small}
\begin{equation}
g_{S}(h(\cdot))\approx g_{T}(\cdot) \;\;\Longrightarrow\;\; \forall J(\cdot),\;\;\frac{\partial J(y_{S})}{\partial x}\propto \frac{\partial J(y_{T})}{\partial x}
\label{eqn:equal}
\end{equation}
\end{small}
where $J(\cdot)$ is an arbitrary function of $y$ that outputs a scalar. $\theta_{h}$ denote parameters of the adapter $h$. Therefore, we use the following distillation loss for \textit{back-distillation}:
\begin{small}
\begin{equation}
\!\underset{\theta_{h}}{\min}\,Loss,\;\; Loss=\mathcal{L}(y_{S},y^{*})+\lambda\cdot\Vert\alpha\frac{\partial J(y_{S})}{\partial x}\!-\!\frac{\partial J(y_{T})}{\partial x}\Vert^2\!\!
\end{equation}
\end{small}
where {\small$\mathcal{L}(y_{S},y^{*})$} is the task loss of the student net; {\small$y^{*}$} denotes the ground-truth label; $\alpha$ is a scaling scalar. This formulation is similar to the Jacobian distillation in \cite{distillJacobian}. We omit {\small$\mathcal{L}(y_{S},y^{*})$}, if we learn the adapter without additional training labels.

\subsection{Learning via back distillation}
\label{sec:backDistill}

It is difficult for most recent techniques, including those for Jacobian distillation\textcolor{red}{\footnotemark[1]}, to directly optimize the above back-distillation loss. We briefly analyze the difficulties as follows. The minimization of the distillation loss is actually to push gradients of the student net towards those of the teacher net. To simplify the notation, we use {\small$D_{S}=\frac{\partial J(y_{S})}{\partial x}$} and {\small$D_{T}=\frac{\partial J(y_{T})}{\partial x}$} to denote gradients \emph{w.r.t.} the feature map in the student net and the teacher net, respectively. As shown in Eqn.~(\ref{eqn:D}), the computation of {\small$D_{S}$} (or {\small$D_{T}$}) is sensitive to feature maps of layers in the {\small$g_{S}$ and $h$} (those in {\small$g_{T}$}). Thus, it requires the student network to yield well-optimized feature maps to enable an effective distillation process. However, it is a chicken-and-egg problem---distilling optimal parameters {\small$\theta_{h}$} and generating optimal feature maps of middle layers: Chaotic initial feature maps hurt the capability of distilling knowledge into {\small$\theta_{h}$}, but the feature maps are produced using {\small$\theta_{h}$}.

To overcome the optimization problem, we need to make gradients of $J$ agnostic with regard to feature maps. Thus, we propose two pseudo-gradients {\small$D'_{S},D'_{T}$} to replace {\small$D_{S},D_{T}$} in the loss, respectively. The pseudo-gradients {\small$D'_{S},D'_{T}$} follow the paradigm in Eqn.~(\ref{eqn:DNew}).
\begin{small}
\begin{subequations}
\begin{align}
D({\bf X},\theta_{h})&\xlongequal[]{\textrm{def}}\left.\frac{\partial J}{\partial x}\right|_{x=x^{(m)}}=G_{y}\frac{\partial y}{\partial x^{(n)}}\cdots\frac{\partial x^{(m+1)}}{\partial x^{(m)}}\label{eqn:D}\\
&=f'_{\textrm{conv}}\circ f'_{\textrm{relu}}\circ f_{\textrm{pool}}^{'\textrm{max}}\circ\cdots\circ f'_{\textrm{conv}}(G_{y})\nonumber\\
D'(\theta_{h})&\xlongequal[]{\textrm{def}}f'_{\textrm{conv}}\circ f'_{\textrm{dummy}}\circ f_{\textrm{pool}}^{'\textrm{avg}}\circ\cdots\circ f'_{\textrm{conv}}\left(G_{y'}\right)\label{eqn:DNew}
\end{align}
\end{subequations}
\end{small}
where we define {\small$G_{y}=\frac{\partial J}{\partial y}$}. Just like in Eqn.~(\ref{eqn:equal}), we assume {\small$g_{S}(h(\cdot))\approx g_{T}(\cdot)\Rightarrow D'_{S}\propto D'_{T}$}. {\small$f'_1\circ f'_2(\cdot)\xlongequal[]{\textrm{def}}f'_1(f'_2(\cdot))$}, each $f'$ is the derivative of the layer function $f$ for back-propagation. {\small${\bf X}$} denotes a set of feature maps of all middle layers, and {\small$x^{(m)}\in{\bf X}$} is the feature map of the $m$-th layer.

In Eqn.~(\ref{eqn:DNew}), we make the following revisions to the computation of gradients, in order to make gradients {\small$D'$} agnostic with regard to {\small${\bf X}$}. We ignore dropout operations and replace derivatives of max-pooling {\small$f_{\textrm{pool}}^{'\textrm{max}}$} with derivatives of average-pooling {\small$f_{\textrm{pool}}^{'\textrm{avg}}$}. We also revise the derivative of the ReLU to either {\small$f^{\textrm{1st}}_{\textrm{dummy}}(\frac{\partial J}{\partial x^{(k)}})\!=\!\frac{\partial J}{\partial x^{(k)}}$} or {\small$f^{\textrm{2nd}}_{\textrm{dummy}}(\frac{\partial J}{\partial x^{(k)}})\!=\!\frac{\partial J}{\partial x^{(k)}}\odot{\bf1}(x_{\textrm{rand}}\!>\!0)$}, where {\small$x_{\textrm{rand}}\!\in\!\mathbb{R}^{s_1\times s_2\times s_3}$} is a random feature map; $\odot$ denotes the element-wise product. For each input image, we set the same random feature map {\small$x_{\textrm{rand}}$} and initial gradients {\small$G_{y'}$} for both {\small$g_{S}$} and {\small$g_{T}$} to make {\small$D'_{S}$} and {\small$D'_{T}$} comparable with each other. Above revisions are made for {\small$D'_{S},D'_{T}$} to ease the back distillation, and they are not related to the computation of the task loss {\small$\mathcal{L}(y,y^{*})$}.



In this way, we conduct the back-distillation algorithm by {\small$\min_{\theta_{h}}Loss\!=\!\mathcal{L}(y_{S},y^{*})\!+\!\lambda\Vert\alpha D'_{S}\!-\!D'_{T}\Vert^2$}. The distillation loss can be optimized by propagating gradients of gradient maps to the upper layers, and we consider this as \textit{back-back-propagation}\textcolor{red}{\footnotemark[1]}. Tables~\ref{tab:exp1} and \ref{tab:exp23} have exhibited the superior performance of the back-back-propagation.

\textbf{Computation of {\small$G_{y'}$}:} According to Eqn.~(\ref{eqn:equal}), $J$ can be any arbitrary function. Thus, we can enumerate functions of $J$ by randomizing different values of {\small$G_{y'}$}. We use each {\small$G_{y'}$} to produce a pair of {\small$D'_{S}$} and {\small$D'_{T}$} for back distillation. For the task of object segmentation, the output is a tensor {\small$y_{S}\in\mathbb{R}^{H\times W\times C}$}, where {\small$H$} and {\small$W$} denote the height and width of the output image, and {\small$C$} indicates the number of segmentation labels. For each image, we randomly sample {\small$G_{y'}\in\mathbb{R}^{H\times W\times C}$}. For the task of single-category classification, the output {\small$y_{S}$} is a scalar. Nevertheless, we can still generate a random matrix {\small$G_{y'}\in\mathbb{R}^{S\times S\times 1}$} ({\small$-1\leq G_{y'}^{ij1}\leq+1$}) for each image ({\small$S\!=\!7$} in experiments), which produces two enlarged pseudo-gradient maps $D'_{S},D'_{T}$ for back distillation\textcolor{red}{\footnotemark[1]}. We normalize {\small$G_{y'}$} to the ranges of {\small$-1\leq G_{y'}^{ijk}\leq+1$}, which ensures a stable distillation in experiments.

\begin{figure}
\centering
\includegraphics[width=0.99\linewidth]{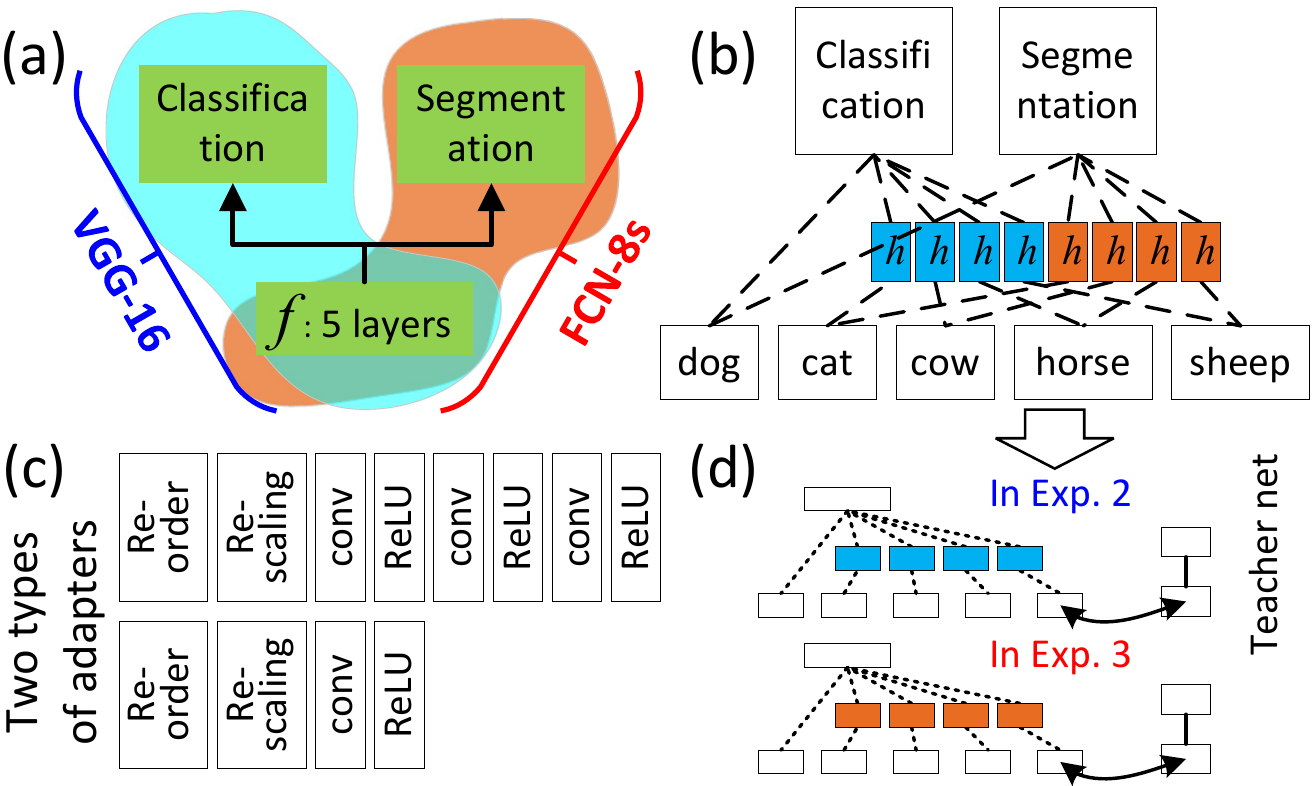}
\vspace{2pt}
\caption{(a) Teacher net; (b) transplant net; (c) two types of adapters; (d) two sequences of transplanting operations.}
\label{fig:exp23}
\end{figure}

\section{Experiments}

To simplify the story, we limit our attention to testing network transplanting operations. We do not discuss other related operations, \emph{e.g.} the fine-tuning of category and task modules and the case in Fig.~\ref{fig:problem}(a), which can be solved via traditional learning strategies.

We designed three experiments to evaluate the proposed method. In Experiment~1, we learned toy transplant nets by inserting adapters between middle layers of pre-trained CNNs. Then, Experiments~2 and 3 were designed considering the real application of learning a transplant net with two task modules (\emph{i.e.} modules for object classification and segmentation) and multiple category modules. As shown in Fig.~\ref{fig:exp23}(b,d), we can divide the entire network-transplanting procedure into an operation sequence of transplanting category modules to the classification module and another operation sequence of transplanting category modules to the segmentation module. Therefore, we separately conducted the two sequences of transplanting operations in Experiments~2 and 3 for more convincing results.

Because our back-distillation strategy decreases the demand for training samples, we tested the learning of adapters with limited numbers of samples (\emph{i.e.} 10, 20, 50, and 100 samples). We even tested network transplanting \textbf{without any} training samples in Experiment 1, \emph{i.e.} optimizing the distillation loss without considering the task loss.


\textbf{Baselines:} We compared our back-distillation method (namely \textit{back-distill}) with two baselines. All baselines exclusively learned the adapter without fine-tuning the task module for fair comparisons. The first baseline only optimized the task loss {\small$\mathcal{L}(y_{S},y^{*})$} without distillation, namely \textit{direct-learn}. The second baseline is the traditional distillation~\cite{distill}, namely \textit{distill}, where the distillation loss is {\small$CrossEntropy(y_{S},y_{T})$}. The distillation was applied to outputs of task modules {\small$g_{S}$} and {\small$g_{T}$}, because except for outputs, other layers in {\small$g_{S}$} and {\small$g_{T}$} did not produce features on similar manifolds. We tested the \textit{distill} method in object segmentation, because unlike single-category classification, segmentation outputs had correlations between soft output labels.

\begin{figure*}[t]
\centering
\includegraphics[width=0.9\linewidth]{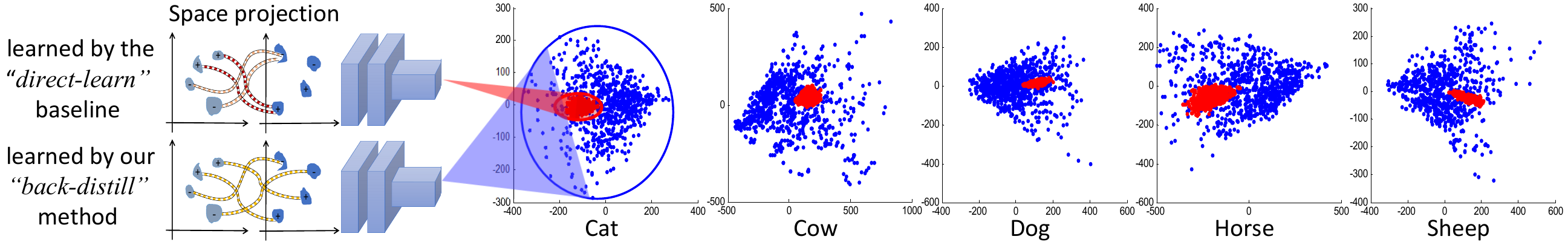}
\vspace{2pt}
\caption{Comparison of the projected feature spaces. For each category, blue points indicate 4096-d \textit{fc8} features of different images, when our method learned the adapter. Red points correspond to \textit{fc8} features of different images, when the adapter was learned by only using the task loss, \emph{i.e.} the \textit{direct-learn} baseline. We visualize the first two principal components of \textit{fc8} features. Because the \textit{direct-learn} baseline usually learned problematic many-to-one projections and projections to forgotten spaces (see Fig.~\ref{fig:space}), most information in $h(x)$ could not pass through ReLU layers to the \textit{fc8} layer. Therefore, given the adapter learned based on the \textit{direct-learn} baseline, units in the \textit{fc8} layer were weakly triggered, and many-to-one projections decreased the diversity of \textit{fc8} features.}
\label{fig:spaceVisual}
\end{figure*}

\textbf{Network structures, datasets, \& details:} We transplanted category modules to a classification module in the first two experiments, and transplanted category modules to a segmentation module in the third experiment. In recent studies, people usually extended the structure of widely-used VGG-16 net~\cite{VGG} to implement classification~\cite{VGG,interpretableCNN} and segmentation~\cite{VGGSeg}, as standard baselines of the two tasks. Thus, as shown in Fig.~\ref{fig:exp23}(a), we can represent a teacher net for both classification and segmentation as a network with a single category module and two task modules. The network branch for classification was exactly a VGG-16 net, and the network branch for segmentation was identical to the FCN-8s model proposed in \cite{VGGSeg}. Because the first five layers of the FCN-8s and those of the VGG-16 share the same structure, we considered the first five layers (including two conv-layers, two ReLU layers, and one pooling layer) as the shared category module and regarded upper layers of the VGG-16 and the FCN-8s as two task modules. Both branches are benchmark networks.

We followed standard experimental settings in \cite{VGGSeg} to learn the FCN-8s for each category, which used the Pascal VOC 2011 dataset (with segmentation labels on 8498 PASCAL images collected by \cite{VOCSeg}). For object classification, we followed standard settings in \cite{interpretableCNN} that used the PASCAL VOC images to learn CNNs for the binary classification of a single category from random images. Note that we only learned and merged teacher networks for five mammal categories, \emph{i.e.} the cat, cow, dog, horse, and sheep categories. Mammal categories share similar object structures, which make features of a category transferable to other categories.

Now, we introduce adapter structures, as shown in Fig.~\ref{fig:exp23}(c). An adapter contained $n$ conv-layers, each followed by a ReLU layer ({\small$n=1$ or $3$} in experiments)\footnote[2]{Each conv-layer in the adapter contained {$M$} filters. Each filter was a {$3\times3\times M$} tensor with a $padding=1$ in Experiment 1 or a {$1\times1\times M$} tensor without padding in Exps.~2 and 3, to avoid changing the size of feature maps, where $M$ is the channel number of $f$.}. In addition, we inserted a ``reorder'' layer and a ``re-scaling'' layer\textcolor{red}{\footnotemark[1]} in front of conv-layers in the adapter. The ``reorder'' layer randomly reordered channels of the features $x$ from the category module, which enlarged the dissimilarity between output features of different category modules. We inserted the ``reorder'' layer to mimic feature states in real applications for fair evaluation. The ``re-scaling'' layer normalized the scale of features $x$ from the category module, \emph{i.e.} {\small$x^{\textrm{out}}=\beta\cdot x$}, for robust network transplanting. {\small$\beta={\mathbb E}_{I\in{\bf I}_{S}}[\Vert f_{S}(I)\Vert_{F}]/{\mathbb E}_{I\in{\bf I}_{T}}[\Vert f(I)\Vert_{F}]$} is a fixed scalar. {\small${\bf I}_{T}$} and {\small${\bf I}_{S}$} denote the image set of the new category and the image set of categories that had been already modeled by the transplant net, respectively\footnote[3]{Because we used the task module in the dog network as the generic task module $g_{S}$, we got ${\bf I}_{S}={\bf I}_{\textrm{dog}}$.}. $f_{S}(I)$ denotes the input feature of $g_{S}$ given an image $I$. $\Vert\cdot\Vert_{F}$ denotes the Frobenius norm. We set the parameter {\small$\alpha={\mathbb E}_{I\in{\bf I}_{T}}[\Vert D'_{T}\Vert]/{\mathbb E}_{I\in{\bf I}_{S}}[\Vert D'_{S}\Vert]$}.

\begin{table*}
\begin{center}
\resizebox{0.95\linewidth}{!}{
\begin{tabular}{c|cl|rrrrr|r|c|cl|rrrrr|r|}
\cline{2-9}\cline{11-18}
\multirow{11}{*}{\begin{sideways}{Insert one conv-layer}\end{sideways}}&
\multicolumn{2}{l|}{\footnotesize{\# of samples}} & cat & cow & dog & horse & sheep & {\bf Avg.} &
\multirow{11}{*}{\begin{sideways}{Insert three conv-layers}\end{sideways}}&
\multicolumn{2}{l|}{\footnotesize{\# of samples}} & cat & cow & dog & horse & sheep & {\bf Avg.}\\
\cline{2-9}\cline{11-18}
&\multirow{2}{*}{100}&{direct-learn} & 12.89
& 3.09
& 12.89
& 10.82
& 9.28
& 9.79
&&
\multirow{2}{*}{100}&{direct-learn} & 9.28
& 6.70
& 12.37
& 11.34
& 3.61
& 8.66\\
&&{back-distill} & {\bf1.55}
& {\bf0.52}
& {\bf3.61}
& {\bf1.55}
& {\bf1.03}
& {\bf1.65}
&&
&{back-distill} & {\bf1.03}
& {\bf2.58}
& {\bf4.12}
& {\bf1.55}
& {\bf2.58}
& {\bf2.37}\\
\cline{2-9}\cline{11-18}
&\multirow{2}{*}{50}&{direct-learn} & 13.92
& 15.98
& 12.37
& 16.49
& 15.46
& 14.84
&&
\multirow{2}{*}{50}&{direct-learn} & 14.43
& 13.92
& 15.46
& 8.76
& 7.22
& 11.96\\
&&{back-distill} & {\bf1.55}
& {\bf0.52}
& {\bf3.61}
& {\bf1.55}
& {\bf1.03}
& {\bf1.65}
&&
&{back-distill} & {\bf3.09}
& {\bf3.09}
& {\bf4.12}
& {\bf2.06}
& {\bf4.64}
& {\bf3.40}\\
\cline{2-9}\cline{11-18}
&\multirow{2}{*}{20}&{direct-learn} & 16.49
& 26.80
& 28.35
& 32.47
& 25.77
& 25.98
&&
\multirow{2}{*}{20}&{direct-learn} & 22.16
& 25.77
& 32.99
& 22.68
& 22.16
& 25.15\\
&&{back-distill} & {\bf1.55}
& {\bf0.52}
& {\bf3.09}
& {\bf1.55}
& {\bf1.03}
& {\bf1.55}
&&
&{back-distill} & {\bf7.22}
& {\bf6.70}
& {\bf7.22}
& {\bf2.58}
& {\bf5.15}
& {\bf5.77}\\
\cline{2-9}\cline{11-18}
&\multirow{2}{*}{10}&{direct-learn} & 39.18
& 39.18
& 35.05
& 41.75
& 38.66
& 38.76
&&
\multirow{2}{*}{10}&{direct-learn} & 36.08
& 32.99
& 31.96
& 34.54
& 34.02
& 33.92\\
&&{back-distill} & {\bf1.55}
& {\bf0.52}
& {\bf3.61}
& {\bf1.55}
& {\bf1.03}
& {\bf1.65}
&&
&{back-distill} & {\bf8.25}
& {\bf15.46}
& {\bf10.31}
& {\bf13.92}
& {\bf10.31}
& {\bf11.65}\\
\cline{2-9}\cline{11-18}
&\multirow{2}{*}{\textcolor{red}{\bf0}}&{direct-learn} & -- & -- & -- & -- & -- & --
&&
\multirow{2}{*}{\textcolor{red}{\bf0}}&{direct-learn} & -- & -- & -- & -- & -- & --\\
&&{back-distill} & {\bf1.55}
& {\bf0.52}
& {\bf4.12}
& {\bf1.55}
& {\bf1.03}
& {\bf1.75}
&&
&{back-distill} & {\bf50.00}
& {\bf50.00}
& {\bf50.00}
& {\bf49.48}
& {\bf50.00}
& {\bf49.90}\\
\cline{2-9}\cline{11-18}
\end{tabular}}
\vspace{2pt}
\caption{Error rates of classification when we insert $n$ conv-layers with ReLU layers to a CNN as the adapter. {\small$n\in\{1,3\}$}. The last row shows the performance of network transplanting without training samples, \emph{i.e.} without optimizing the task loss {\small$\mathcal{L}(y_{S},y^{*})$}.}
\label{tab:exp1}
\end{center}
\end{table*}

\subsection{Exp. 1: Adding adapters to pre-trained CNNs}

In this experiment, we conducted a toy test, \emph{i.e.} inserting and learning an adapter between a category module and a task module to test network transplanting. Here, we only considered networks with VGG-16 structures~\cite{VGG} for single-category classification. These networks were strongly supervised using all training samples and achieved error rates of 1.6\%, 0.6\%, 4.1\%, 1.6\%, and 1.0\% for the classification of the cat, cow, dog, horse, and sheep categories, respectively. Then, we learned two types of adapters (see Fig.~\ref{fig:exp23}(c)), which have been introduced before.

Because the classification output $y'$ is a scalar without neighboring outputs to provide correlations, we simply set {\small$G_y'\!=\!1$} without any gradient randomization. Instead, we used the revised dummy ReLU operations in Eqn.~(\ref{eqn:DNew}) to ensure the value diversity of {\small$D'_{S},D'_{T}$} for learning. More specifically, we used {\small$f'_{\textrm{dummy}}=f^{\textrm{2nd}}_{\textrm{dummy}}$} to compute expedient derivatives of ReLU operations in the task module, and used {\small$f'_{\textrm{dummy}}=f^{\textrm{1st}}_{\textrm{dummy}}$} in the adapter\footnote[4]{All derivative functions in Eqn.~(\ref{eqn:DNew}) are only used for distillation, which will not affect gradient propagations from $\mathcal{L}(y,y^{*})$.}. We set {\small$\lambda=10.0/{\mathbb E}_{I\in{\bf I}_{T}}[D'_{T}]$} for object classification in Experiments 1 and 2.

In Fig.~\ref{fig:spaceVisual}, we compared the space of \textit{fc8} features, when we used our method and the \textit{direct-learn} baseline, respectively, to learn the adapter with three conv-layers. The adapter learned by our method passed much stronger information to the final \textit{fc8} layer and yielded more diverse features. It demonstrates that our method better avoided problematic projections in Fig.~\ref{fig:space} than the \textit{direct-learn} baseline.

\begin{table*}
\begin{center}
\resizebox{0.9\linewidth}{!}{\begin{tabular}{cl|rrrrr||cl|rrrrr}
\multicolumn{14}{c}{Experiment 2: transplanting to a classification module}\\
\hline
\multicolumn{2}{l|}{\footnotesize{$\!\!\!\!\!$\# of samples}} & cat & cow & horse & sheep & {\bf Avg.} & \multicolumn{2}{l|}{\footnotesize{\# of samples}} & cat & cow & horse & sheep & {\bf Avg.}\\
\hline
\multirow{2}{*}{100}&{direct-learn}& 20.10
& 12.37
& 18.56
& 11.86
& 15.72
&\multirow{2}{*}{20}&{direct-learn}& 31.96
& 37.11
& 39.69
& 35.57
& 36.08\\
&{back-distill} & {\bf9.79}
& {\bf5.67}
& {\bf8.25}
& {\bf4.64}
& {\bf7.09}
&&{back-distill} & {\bf21.13}
& {\bf35.57}
& {\bf32.47}
& {\bf22.68}
& {\bf27.96}\\
\hline
\multirow{2}{*}{50}&{direct-learn}& 22.68
& 19.59
& 19.07
& 14.95
& 19.07
&\multirow{2}{*}{10}&{direct-learn}& 41.75
& {\bf37.63}
& {\bf44.33}
& 33.51
& 39.31\\
&{back-distill} & {\bf10.82}
& {\bf18.04}
& {\bf13.92}
& {\bf5.15}
& {\bf11.98}
&&{back-distill} & {\bf34.02}
& 42.27
& 44.85
& {\bf33.51}
& {\bf38.66}\\
\hline
\end{tabular}}
\resizebox{0.9\linewidth}{!}{\begin{tabular}{cl|rrrrr||cl|rrrrr}
\multicolumn{14}{c}{Experiment 3: transplanting to a segmentation module}\\
\hline
\multicolumn{2}{l|}{\footnotesize{$\!\!\!\!\!$\# of samples}} & cat & cow & horse & sheep & {\bf Avg.} & \multicolumn{2}{l|}{\footnotesize{\# of samples}} & cat & cow & horse & sheep & {\bf Avg.}\\
\hline
\multirow{3}{*}{100}&{direct-learn}& 76.54
& 74.60
& 81.00
& 78.37
& 77.63
&\multirow{3}{*}{20}&{direct-learn}& 71.13
& 74.82
& 76.83
& 77.81
& 75.15\\
&{distill} &74.65
& 80.18
& 78.05
& 80.50
& 78.35
&&{distill} &71.17
& 74.82
& 76.05
& 78.10
& 75.04\\
&{back-distill} & {\bf85.17}
& {\bf90.04}
& {\bf90.13}
& {\bf86.53}
& {\bf87.97}
&&{back-distill} & {\bf84.03}
& {\bf88.37}
& {\bf89.22}
& {\bf85.01}
& {\bf86.66}\\
\hline
\multirow{3}{*}{50}&{direct-learn}& 71.30
& 74.76
& 76.83
& 78.47
& 75.34
&\multirow{3}{*}{10}&{direct-learn}& 70.46
& 74.74
& 76.49
& 78.25
& 74.99\\
&{distill} &68.32
& 76.50
& 78.58
& 80.62
& 76.01
&&{distill} &70.47
& 74.74
& 76.83
& 78.32
& 75.09\\
&{back-distill} & {\bf83.14}
& {\bf90.02}
& {\bf90.46}
& {\bf85.58}
& {\bf87.30}
&&{back-distill} & {\bf82.32}
& {\bf89.49}
& {\bf85.97}
& {\bf83.50}
& {\bf85.32}\\
\hline
\end{tabular}}
\vspace{2pt}
\caption{(top) Error rate of single-category classification when we transplanted the classification module from a pre-trained \textit{dog} network to the network of the target category in Experiment 2. The adapter contained three conv-layers. (bottom) Pixel accuracy of object segmentation when we transplanted the segmentation module from a \textit{dog} network to the network of the target category in Experiment 3. The adapter contained a conv-layer and a ReLU layer.}
\label{tab:exp23}
\end{center}
\end{table*}

In Table \ref{tab:exp1}, we compared our \textit{back-distill} method with the \textit{direct-learn} baseline when we inserted an adapter with a single conv-layer and when we inserted an adapter with three conv-layers. Table~\ref{tab:exp1}(left) shows that compared to the $9.79\%$--$38.76\%$ error rates of the \textit{direct-learn} baseline, our \textit{back-distill} method yielded a significant lower classification error ($1.55\%$--$1.75\%$). \textit{Even without any training samples, our method still outperformed the \textit{direct-learn} method with 100 training samples.}

Note that given an adapter with multiple conv-layers (\emph{e.g.} three conv-layers) without any training samples, our \textit{back-distill} method was not powerful enough to learn the adapter. Because deeper adapters with more parameters had more flexibility in representation, which required stronger supervision to avoid over-fitting. For example, in the last row of Table~\ref{tab:exp1}, our method successfully optimized an adapter with a single conv-layer (the error rate was 1.75\%), but was hampered when the adapter had three conv-layers (the error rate was 49.9\%, and our method produced a biased short-cut solution).

\subsection{Exp. 2: Operation sequences of transplanting category modules to the classification module}

In this experiment, we evaluated the performance of transplanting category modules to the classification module. We considered the classification module of the dog\footnote[5]{Because the dog category contained more training samples, the CNN for the dog was believed to be better learned. Thus, we used the task module learned for dog images as a generic task module.} as a generic one. We transplanted category modules of other four mammal categories to this task module. According to the experience in Experiment 1, we set the adapter to contain three conv-layers. Following Eqn.~(\ref{eqn:DNew}), we used the {\small$f'_{\textrm{dummy}}=f^{\textrm{1st}}_{\textrm{dummy}}$} operation to compute derivatives of ReLU operations in both the task module and the adapter\textcolor{red}{\footnotemark[3]}. The only exception was the lowest ReLU operation of the task module, for which we applied {\small$f'_{\textrm{dummy}}=f^{\textrm{2nd}}_{\textrm{dummy}}|_{x_{\textrm{rand}}}$}. We generated {\small$x_{\textrm{rand}}=[x',x',\ldots,x']$} for each input image by concatenating {\small$s_3$} matrices $x'$ along the third dimension, where {\small$x'\in\mathbb{R}^{s_1\times s_2\times 1}$} contained 20\%/80\% positive/negative elements. The generation of $G_{y'}$ is introduced in Section~\ref{sec:backDistill}.

Table~\ref{tab:exp23} shows the performance when we transplanted the category module to a task module oriented to categories with similar structures. We tested our method with a few (10--100) training samples. When there were more than 50 training samples, our method yielded about a half classification error of the \textit{direct-learn} baseline.

\begin{table*}
\begin{center}
\resizebox{0.95\linewidth}{!}{\begin{tabular}{c|cl|rrrrr||cl|rrrrr}
\hline
\multicolumn{3}{c|}{\footnotesize{$\!\!\!\!\!$\# of samples}} & cat & cow & horse & sheep & {\bf Avg.} & \multicolumn{2}{l|}{\footnotesize{\# of samples}} & cat & cow & horse & sheep & {\bf Avg.}\\
\hline
\multirow{4}{*}{\begin{sideways}{{\footnotesize Classification}}\end{sideways}}&\multirow{2}{*}{100}&{direct-learn}&14.43
&20.62
&17.01
&11.86
&15.98
&\multirow{2}{*}{20}&{direct-learn}&25.26
&24.23
&39.18
&23.71
&28.10
\\
&&{back-distill} &{\bf5.67}
&{\bf3.61}
&{\bf6.70}
&{\bf2.58}
&{\bf4.64}
&&{back-distill} &{\bf17.01}
&{\bf19.59}
&{\bf23.71}
&{\bf14.95}
&{\bf18.82}
\\
\cline{2-15}
&\multirow{2}{*}{50}&{direct-learn}&21.13
&23.71
&15.46
&10.31
&17.65
&\multirow{2}{*}{10}&{direct-learn}&42.27
&36.60
&40.72
&39.18
&39.69
\\
&&{back-distill} &{\bf7.22}
&{\bf9.28}
&{\bf8.76}
&{\bf5.67}
&{\bf7.73}
&&{back-distill} &{\bf42.27}
&{\bf32.99}
&{\bf28.35}
&{\bf30.41}
&{\bf33.51}
\\
\hline
\hline
\multirow{2}{*}{\begin{sideways}{{$\textrm{Segmen}\atop\textrm{tation}$}}\end{sideways}}&\multirow{2}{*}{10}&{direct-learn}&64.97
&69.65
&80.26
&69.87
&71.19
&\multirow{2}{*}{20}&{direct-learn}&68.69
&81.02
&71.88
&72.65
&73.56
\\
&&{back-distill} &{\bf74.59}
&{\bf83.51}
&{\bf82.08}
&{\bf80.21}
&{\bf80.10}
&&{back-distill} &{\bf73.34}
&{\bf84.78}
&{\bf81.40}
&{\bf81.04}
&{\bf80.14}
\\
\hline
\end{tabular}}
\vspace{2pt}
\caption{(top) Error rate of single-category classification when the classification module was learned for both mammals and dissimilar categories. (bottom) Pixel accuracy of object segmentation, the segmentation module was learned for both mammals and dissimilar categories. Other experimental settings were the same as in Experiments~2 and 3. We did not show the result of the dog category, because we needed to compare average performance in this table with results in Table~\ref{tab:exp23}.}
\label{tab:dissimilar}
\end{center}
\end{table*}

\subsection{Exp. 3: Operation sequences of transplanting category modules to the segmentation module}

In this experiment, we evaluated the performance of transplanting category modules to the segmentation module. Five FCNs were strongly supervised using all training samples for single-category segmentation. These networks achieved pixel-level segmentation accuracies (defined in \cite{VGGSeg}) of 95.0\%, 94.7\%, 95.8\%, 94.6\%, and 95.6\% for the cat, cow, dog, horse, and sheep categories, respectively. Like in Experiment 2, we considered the segmentation module of the dog\textcolor{red}{\footnotemark[5]} as a generic one. We transplanted category modules of other four mammal categories to this task module.
According to the experience in Experiment 1, we set the adapter to contain one conv-layer. Following Eqn.~(\ref{eqn:DNew}), we used the {\small$f'_{\textrm{dummy}}=f^{\textrm{1st}}_{\textrm{dummy}}$} operation to compute derivatives of ReLU operations\textcolor{red}{\footnotemark[3]}. We set {\small$\lambda=1.0/{\mathbb E}_{I}[D'_{T}]$} for all categories in this experiment.

Table~\ref{tab:exp23} compares pixel-level segmentation accuracy between our method and the \textit{direct-learn} baseline. We tested our method with a few (10--100) training samples. Our method exhibited 10\%--12\% higher accuracy than the \textit{direct-learn} baseline.

\subsection{Transplant to similar or dissimilar categories?}

Theoretically, just like transfer learning, the task module should deal with a set of categories that have similar structures with the new category to ensure a high efficiency. In fact, identifying categories with similar structures is still an open problem. People manually define sets of similar categories, \emph{e.g.} learning a task module for mammals and learning another task module for different vehicles.

In order to quantitatively evaluate the performance of transplanting to dissimilar categories, we designed new task modules for additional testing. We considered the first four categories of the VOC dataset, \emph{i.e.} aeroplane, bicycle, bird, and boat, to have dissimilar structures with mammals. In Experiment~2/Experiment~3, for the transplanting of a mammal category (let us take the cat for example) to a classification/segmentation module, we learned a ``leave-one-out'' classification/segmentation module to deal with all four dissimilar categories and all mammal categories except the cat.

Table~\ref{tab:dissimilar} shows the performance of transplanting to a task module trained for both similar and dissimilar categories. Our method outperformed the baseline. Compared to the performance of transplanting to a task module oriented to similar categories in Table~\ref{tab:exp23}, transplanting to a more generic task modules for dissimilar categories hurt the segmentation performance but boosted the classification performance. It is because forcing a task module to handle dissimilar categories sometimes made the task module encode more generic and robust representations, while it may also let the task module ignore details of mammal categories.

\section{Conclusions and discussion}

In this paper, we focused on a new task, \emph{i.e.} merging pre-trained teacher nets into a generic, modular transplant net with a few or even without training annotations. We discussed the importance and core challenges of this task. We developed the back-distillation algorithm as a theoretical solution to the challenging space-projection problem.

The back-distillation strategy significantly decreases the demand for training samples. Experimental results demonstrated the superior efficiency of our method. Our method without any training samples even outperformed the baseline with 100 training samples, as shown in Table~\ref{tab:exp1}(left).



The growth of a large transplant net for different categories and tasks can be divided into lots of elementary operations of network transplanting (see Fig.~\ref{fig:problem} and supplementary materials for more discussions). When the transplant net has been learned, we can optionally fine-tune task modules using training samples of multiple categories. Note that unlike the back distillation, the performance of fine-tuning depends on the number of training samples. Thus, given a few samples, whether an additional fine-tuning will increase or decrease the generality of the transplant net is a difficult question, and it requires sophisticated analysis in the future.

{\small
\bibliographystyle{ieee}
\bibliography{TheBib}

\begin{thebibliography}{10}\itemsep=-1pt

\bibitem{meta2}
M.~Andrychowicz, M.~Denil, S.~G. Colmenarejo, M.~W. Hoffman, D.~Pfau,
  T.~Schaul, B.~Shillingford, and N.~de~Freitas.
\newblock Learning to learn by gradient descent by gradient descent.
\newblock {\em In NIPS}, 2016.

\bibitem{distillDistributed}
R.~Anil, G.~Pereyra, A.~Passos, R.~Ormandi, G.~E. Dahl, and G.~E. Hinton.
\newblock Large scale distributed neural network training through online
  distillation.
\newblock {\em In ICLR}, 2018.

\bibitem{infoGAN}
X.~Chen, Y.~Duan, R.~Houthooft, J.~Schulman, I.~Sutskever, and P.~Abbeel.
\newblock Infogan: Interpretable representation learning by information
  maximizing generative adversarial nets.
\newblock {\em In NIPS}, 2016.

\bibitem{meta1}
Y.~Chen, M.~W. Hoffman, S.~G. Colmenarejo, M.~Denil, T.~P. Lillicrap,
  M.~Botvinick, and N.~de~Freitas.
\newblock Learning to learn without gradient descent by gradient descent.
\newblock {\em In ICML}, 2017.

\bibitem{netMerger}
Y.-M. Chou, Y.-M. Chan, J.-H. Lee, C.-Y. Chiu, and C.-S. Chen.
\newblock Unifying and merging well-trained deep neural networks for inference
  stage.
\newblock {\em In arXiv:1805.04980}, 2018.

\bibitem{PathNet}
C.~Fernando, D.~Banarse, C.~Blundell, Y.~Zwols, D.~Hay, A.~A. Rusu, A.~Pritzel,
  and D.~Wierstra.
\newblock Pathnet: Evolution channels gradient descent in super neural
  networks.
\newblock {\em In arXiv:1701.08734}, 2017.

\bibitem{distillDecisionTree}
N.~Frosst and G.~Hinton.
\newblock Distilling a neural network into a soft decision tree.
\newblock {\em In arXiv:1711.09784}, 2017.

\bibitem{UnsuperTransferCNN}
Y.~Ganin and V.~Lempitsky.
\newblock Unsupervised domain adaptation in backpropagation.
\newblock {\em In {ICML}}, 2015.

\bibitem{VOCSeg}
B.~Hariharan, P.~Arbelaez, L.~Bourdev, S.~Maji, and J.~Malik.
\newblock Semantic contours from inverse detectors.
\newblock {\em In ICCV}, 2011.

\bibitem{betaVAE}
I.~Higgins, L.~Matthey, A.~Pal, C.~Burgess, X.~Glorot, M.~Botvinick,
  S.~Mohamed, and A.~Lerchner.
\newblock $\beta$-vae: learning basic visual concepts with a constrained
  variational framework.
\newblock {\em In ICLR}, 2017.

\bibitem{distill}
G.~Hinton, O.~Vinyals, and J.~Dean.
\newblock Distilling the knowledge in a neural network.
\newblock {\em In NIPS Workshop}, 2014.

\bibitem{deepForest1}
P.~Kontschieder, M.~Fiterau, A.~Criminisi, and S.~R. Bul\`{o}.
\newblock Deep neural decision forests.
\newblock {\em In ICCV}, 2015.

\bibitem{meta3}
K.~Li and J.~Malik.
\newblock Learning to optimize.
\newblock {\em In arXiv:1606.01885}, 2016.

\bibitem{LearnNetStruct2}
C.~Liu, B.~Zoph, J.~Shlens, W.~Hua, L.-J. Li, L.~Fei-Fei, A.~Yuille, J.~Huang,
  and K.~Murphy.
\newblock Progressive neural architecture search.
\newblock {\em In arXiv:1712.00559}, 2017.

\bibitem{VGGSeg}
J.~Long, E.~Shelhamer, and T.~Darrel.
\newblock Fully convolutional networks for semantic segmentation.
\newblock {\em In CVPR}, 2015.

\bibitem{ProgressiveNN}
A.~A. Rusu, N.~C. Rabinowitz, G.~Desjardins, H.~Soyer, J.~Kirkpatrick,
  K.~Kavukcuoglu, R.~Pascanu, and R.~Hadsell.
\newblock Progressive neural networks.
\newblock {\em In arXiv:1606.04671}, 2016.

\bibitem{capsule}
S.~Sabour, N.~Frosst, and G.~E. Hinton.
\newblock Dynamic routing between capsules.
\newblock {\em In NIPS}, 2017.

\bibitem{InformationBottleneck2}
R.~Schwartz-Ziv and N.~Tishby.
\newblock Opening the black box of deep neural networks via information.
\newblock {\em In arXiv:1703.00810}, 2017.

\bibitem{continualLearning}
J.~Schwarz, J.~Luketina, W.~M. Czarnecki, A.~Grabska-Barwinska, Y.~W. Teh,
  R.~Pascanu, and R.~Hadsell.
\newblock Progress \& compress: A scalable framework for continual learning.
\newblock {\em In arXiv:1805.06370}, 2018.

\bibitem{NetMixOfExperts}
N.~Shazeer, A.~Mirhoseini, K.~Maziarz, A.~Davis, Q.~Le, G.~Hinton, and J.~Dean.
\newblock outrageously large neural networks: the sparsely-gated
  mixture-of-experts layer.
\newblock {\em In ICLR}, 2017.

\bibitem{VGG}
K.~Simonyan and A.~Zisserman.
\newblock Very deep convolutional networks for large-scale image recognition.
\newblock {\em In {ICLR}}, 2015.

\bibitem{distillJacobian}
S.~Srinivas and F.~Fleuret.
\newblock Knowledge transfer with jacobian matching.
\newblock {\em In ICML}, 2018.

\bibitem{metaContinual}
R.~Vuorio, D.-Y. Cho, D.~Kim, and J.~Kim.
\newblock Meta continual learning.
\newblock {\em In arXiv:1806.06928}, 2018.

\bibitem{InformationBottleneck}
N.~Wolchover.
\newblock New theory cracks open the black box of deep learning.
\newblock {\em In Quanta Magazine}, 2017.

\bibitem{RNNTree}
M.~Wu, M.~C. Hughes, S.~Parbhoo, M.~Zazzi, V.~Roth, and F.~Doshi-Velez.
\newblock Beyond sparsity: Tree regularization of deep models for
  interpretability.
\newblock {\em In NIPS TIML Workshop}, 2017.

\bibitem{InterRCNN}
T.~Wu, X.~Li, X.~Song, W.~Sun, L.~Dong, and B.~Li.
\newblock Interpretable r-cnn.
\newblock {\em In arXiv:1711.05226}, 2017.

\bibitem{LifelongLearning}
J.~Yoon, E.~Yang, J.~Lee, and S.~J. Hwang.
\newblock Lifelong learning with dynamically expandable networks.
\newblock {\em In ICLR}, 2018.

\bibitem{CNNAnalysis_2}
J.~Yosinski, J.~Clune, Y.~Bengio, and H.~Lipson.
\newblock How transferable are features in deep neural networks?
\newblock {\em In {NIPS}}, 2014.

\bibitem{distillAttention}
S.~Zagoruyko and N.~Komodakis.
\newblock Paying more attention to attention: improving the performance of
  convolutional neural networks via attention transfer.
\newblock {\em In arXiv:1612.03928}, 2017.

\bibitem{interpretableCNN}
Q.~Zhang, Y.~N. Wu, and S.-C. Zhu.
\newblock Interpretable convolutional neural networks.
\newblock {\em In CVPR}, 2018.

\bibitem{LearnNetStruct4}
Z.~Zhong, J.~Yan, and C.-L. Liu.
\newblock Practical network blocks design with q-learning.
\newblock {\em In AAAI}, 2018.

\bibitem{deepForest2}
Z.-H. Zhou and J.~Feng.
\newblock Deep forest: Towards an alternative to deep neural networks.
\newblock {\em In IJCAI}, 2017.

\bibitem{LearnNetStruct1}
B.~Zoph and Q.~V. Le.
\newblock Neural architecture search with reinforcement learning.
\newblock {\em In ICLR}, 2017.

\bibitem{LearnNetStruct3}
B.~Zoph, V.~Vasudevan, J.~Shlens, and Q.~V. Le.
\newblock Learning transferable architectures for scalable image recognition.
\newblock {\em In arXiv:1707.07012}, 2017.

\end{thebibliography}
}

\newpage
\onecolumn
\appendix
\section*{Computation of gradients \emph{w.r.t.} the distillation loss}

In order to optimize the distillation loss, we need to compute $\frac{\partial D'_{S}(\theta_{h})}{\partial\theta_{h}}$, \emph{i.e.} $\frac{\partial D'(\theta)}{\partial\theta}$ with respect to the following $D'(\theta)$.
\begin{equation}
\quad D'(\theta)\xlongequal[]{\textrm{def}}G_{y'}\frac{\partial y'}{\partial x^{(n)}}\cdots\left.\frac{\partial x^{(m+1)}}{\partial x^{(m)}}\right|_{x=x^{(m)}}
=f'_{\textrm{conv}}\circ f'_{\textrm{dummy}}\circ f_{\textrm{pool}}^{'\textrm{avg}}\circ\cdots\circ f'_{\textrm{conv}}\left(G_{y'}\right)\nonumber
\end{equation}

Thus, we first explore the close-form formulation of the function $D'(\theta)$. As shown in the above equation, we can transform the back-propagation process for computing $D'(\theta)$ as a number of cascaded functions $f'_{\textrm{conv}},\ldots,f_{\textrm{pool}}^{'\textrm{avg}},f'_{\textrm{dummy}},f'_{\textrm{conv}}$, which are derivatives of $f_{\textrm{conv}},\ldots,f_{\textrm{pool}}^{\textrm{avg}},f_{\textrm{dummy}},f_{\textrm{conv}}$. Since we have formulated $f'_{\textrm{dummy}}$ in the manuscript and it is easy to obtain $f_{\textrm{pool}}^{'\textrm{avg}}$, we mainly focus on the formulation of $f'_{\textrm{conv}}$.

In general, it is not difficult to derive the derivative of any convolution operation. Here, we focus on the most common case, \emph{i.e.} the convolution operation with a padding $\pm p$ and a stride of 1. Given a tensor $x\in{\mathbb R}^{M\times M\times D}$ and $C$ convolutional filter with weights $w\in{\mathbb R}^{m\times m\times D\times C}$ and a bias term $b\in{\mathbb R}^{C}$, the convolution can be written as $y=x\otimes w+b$. For VGG networks, people usually set $m=2p+1$. $w$ can further absorb $b$ by adding the $(D+1)-th$ channel to $w$ and adding the $(D+1)-th$ channel to $x$ with $x_{:,:,D+1,:}=1$. Thus, we can obtain
\begin{equation}
\begin{split}
y&=x\otimes w\\
\frac{\partial J}{\partial x}&=\frac{\partial J}{\partial y}\otimes W
\end{split}
\nonumber
\end{equation}
where $W\in{\mathbb R}^{m\times m\times C\times D}$ and $W_{i,j,k,l}=w_{m+1-i,m+1-j,l,k}$. Thus, we can write the derivative of $f_{\textrm{conv}}$ as
\begin{equation}
f'_{\textrm{conv}}(G')=G'\otimes W
\nonumber
\end{equation}
In this way, we obtain the close-form formulation of the function $D'(\theta)$. We can easily compute $\frac{\partial D'(\theta)}{\partial\theta}$ using the chain rule of back propagation. We can consider this process as a \textit{back-back-propagation}.

\subsection*{About the case of using an enlarged $G_{y'}$}

When we use an enlarged pseudo-gradient {\small$G_{y'}\in [-1,+1]^{S\times S\times 1}$}, we can obtain an enlarged gradient map $D'_{S}$. As discussed above, the computation of $D'_{S}$ can be considered as a number of cascaded functions $f'_{\textrm{conv}},\ldots,f_{\textrm{pool}}^{'\textrm{avg}},f'_{\textrm{dummy}},f'_{\textrm{conv}}$ with the input $G_{y'}$, which are quite similar to the forward propagation in the neural network.

\section*{Visualization of network structures used in three experiments}

\vspace{10pt}
\includegraphics[width=0.7\linewidth]{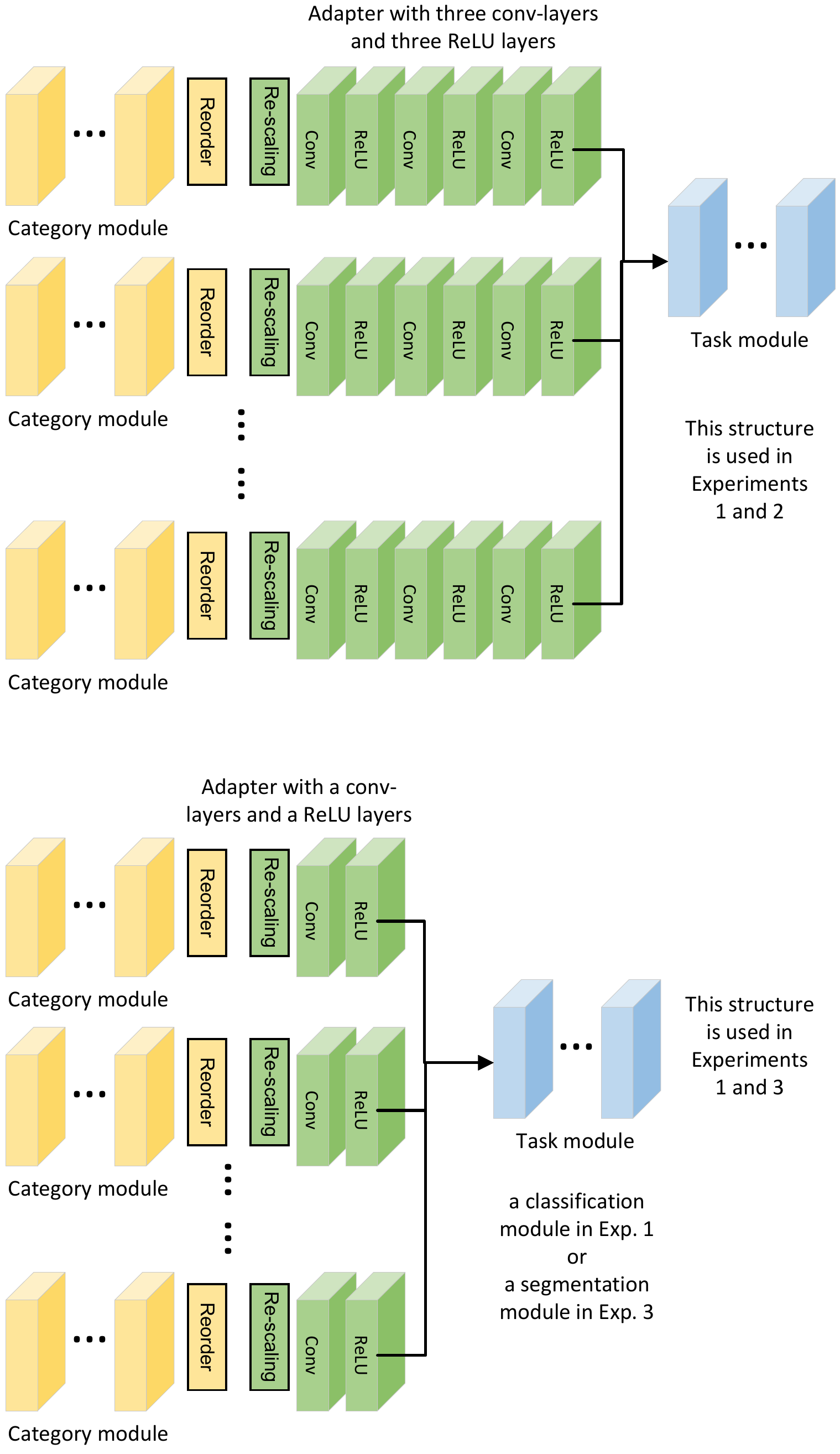}

\newpage
\section*{How to learn a large transplant network}

In the paper, we limit our attention to the core technique of back distillation for network transplanting and do preliminary experiments to demonstrate the effectiveness of the proposed method. Here, we would like to explain how to use the proposed back-distillation algorithm to gradually grow a large transplant network. The basic idea of growing a large transplant network has been shown in Figure 1 of the paper. We can divide the complex procedure of building a large transplant network into elementary operations of network transplanting.

Generally speaking, there are two types of operations during the learning of the transplant network, \emph{i.e.} adding a new task module and adding a new category module.

\textbf{Adding a new task module:} Compared to adding category modules, adding task modules is relatively easier. There are two ways to add a task module.

Firstly, given a single or multiple pre-trained category modules and training samples of the category/categories, we can directly learn a new task module upon features of the category module(s). This learning process has been shown as ``Step 1'' in Figure 1, which follows traditional learning strategy without network transplanting.

Secondly, when the specific network is pre-trained with a single category module $f$ and multiple task modules, we can transplant the category module $f$ to a generic task module $g_{S}$ in the target transplant network. In this case, task modules (those do not correspond to $g_{S}$) will automatically be added to the transplant network, although these task modules only connect with a single category module $f$. This case has been shown as ``Step 2'' in Figure 1, in which the Task-3 module is added to the transplant network when we connect the category-$D$ module to the Task-2 module of the transplant network.

\textbf{Adding a new category module:} We can also divide the insertion of a category modules into the following two cases. The first case is the traditional network transplanting that is introduced in this paper. The second case is to build more connections between category modules and task modules that have already been contained by the transplant network. This case is shown as ``Step 3'' in Figure 1. We learn new adapters to connect existing category modules and task modules via network transplanting.


\section*{Comparing our back-distillation with the Jacobian distillation}

The loss for the Jacobian distillation~\cite{distillJacobian} is similar to Equation (3). There are two essential differences between them, which make the Jacobian distillation not applicable to network transplanting. (i) The Jacobian distillation does not balance magnitudes of neural activations between the category module and the task module, which significantly increases the difficulties of network transplanting. (ii) More crucially, the Jacobian distillation uses real gradients of the network instead of using pseudo-gradients. However, when we fixed the upper task module during network transplanting, the task module usually blocks most signals during the forward propagation. Thus, during the back propagation, real gradients usually cannot pass ReLU layers in the task module to produce informative Jacobians for distillation.

\section*{The core challenge}

To clarify the challenge of learning the adapter, we will compare the following two cases, \emph{i.e.} 1) learning both the adapter $h$ and the task module $g_{S}$ and 2) learning the adapter $h$ but fixing the task module $g_{S}$. The traditional method of directly optimizing the task loss can easily solve the first case but will be hampered in the second case.

\textbf{Case 1, Learning both the adapter $h$ and the task module $g_{S}$:} This case corresponds to the traditional problem of deep learning. We initialize the task module $g_{S}$ with random parameters, so in the first epoch of learning, input features $f$ will produce random activations in layers of both $h$ and $g_{S}$, and positive neural activations will pass through ReLU layers to make random predictions $y_{S}$. Successfully passing information to the final output $y_{S}$ is quite important, because we can obtain gradients of the task loss to optimize $h$ and $g_{S}$. In this way, both $h$ and $g_{S}$ can be well learned.

\textbf{Case 2, learning the adapter $h$ but fixing the task module $g_{S}$:} Unlike Case 1, we use a pre-trained task module $g_{S}$, and we fix its parameters during the learning process. We only initialize the adapter $h$ with random parameters. However, as shown in Fig. 3(left) in the paper, the task module $g_{S}$ has vast forgotten space of its input feature, and $g_{S}$ can only pass very specific features to the final output. In other words, $g_{S}$ cannot pass most information of $h$'s features to the final output $y_{S}$ in early epochs. Thus, we will not obtain informative gradients to optimize parameters in $h$.

\end{document}